\documentclass[]{spie}  

\usepackage{amsmath,amsfonts,amssymb}
\usepackage{graphicx}
\usepackage[colorlinks=true, allcolors=blue]{hyperref}

\title{The Mertens Unrolled Network (MU-Net): A High Dynamic Range Fusion Neural Network for Through the Windshield Driver Recognition}

\author[a]{Max Ruby}
\author[b]{David S. Bolme}
\author[b]{Joel Brogan}
\author[b]{David Cornett III}
\author[c]{Baldemar Delgado}
\author[d]{Gavin Jager}
\author[b]{Christi Johnson}
\author[c]{Jose Martinez-Mendoza}
\author[b]{Hector Santos-Villalobos}
\author[b]{Nisha Srinivas}

\affil[a]{Purdue University, 610 Purdue Mall, West Lafayette, USA}
\affil[b]{Oak Ridge National Laboratory, 1 Bethel Valley Rd, Oak Ridge, USA}
\affil[c]{Texas A\&M University-Kingsville, 700 University Blvd, Kingsville, USA}
\affil[d]{University of Michigan, 500 S. State St, Ann Arbor, USA}

\authorinfo{Further author information: (Send correspondence to Hector Santos-Villalobos)\\Hector Santos-Villalobos: E-mail: hsantos@ornl.gov, Telephone: 1 865 574 0215\\  David S. Bolme: E-mail: bolmeds@ornl.gov, Telephone: 1 865 576 0300\\ \\Notice:  This manuscript has been authored by UT-Battelle, LLC, under contract DE-AC05-00OR22725 with the US Department of Energy (DOE). The US government retains and the publisher, by accepting the article for publication, acknowledges that the US government retains a nonexclusive, paid-up, irrevocable, worldwide license to publish or reproduce the published form of this manuscript, or allow others to do so, for US government purposes. DOE will provide public access to these results of federally sponsored research in accordance with the DOE Public Access Plan (http://energy.gov/downloads/doe-public-access-plan).}

\begin{document} 
\maketitle

\begin{abstract}
      Face recognition of vehicle occupants through windshields in unconstrained environments poses a number of unique challenges ranging from glare, poor illumination, driver pose and motion blur. In this paper, we further develop the hardware and software components of a custom vehicle imaging system to better overcome these challenges. After the build out of a physical prototype system that performs High Dynamic Range (HDR) imaging, we collect a small dataset of through-windshield image captures of known drivers. We then re-formulate the classical Mertens-Kautz-Van Reeth HDR fusion algorithm as a pre-initialized neural network, which we name the Mertens Unrolled Network (MU-Net), for the purpose of fine-tuning the HDR output of through-windshield images. Reconstructed faces from this novel HDR method are then evaluated and compared against other traditional and experimental HDR methods in a pre-trained state-of-the-art (SOTA) facial recognition pipeline, verifying the efficacy of our approach. 
\end{abstract}

\keywords{MU-Net, HDR, Facial Recognition, Through Windshield}

\section{INTRODUCTION}
\label{sec:intro}  

Face recognition of vehicle occupants in unconstrained environments, particularly through the windshield, poses a number of challenges. Previous research has shown that artifacts introduced by a windshield can greatly impact a camera's ability to image within the vehicle interior \cite{bhise2008effect}. 

Additionally, images captured in this scenario are typically from long distances and moderate speeds, which reduces the amount of light available to the sensor. Low-intensity ultraviolet or near-infrared (NIR) illumination has become increasingly ineffective in alleviating these issues as tinted windshields that block these wavelengths gain popularity. Likewise, increasing the exposure time cannot fix the problems associated with that since the vehicle is moving, and motion blur artifacts caused by increasing exposure time significantly degrades face recognition quality.

\begin{figure}
\centering
\includegraphics[width=5.5in]{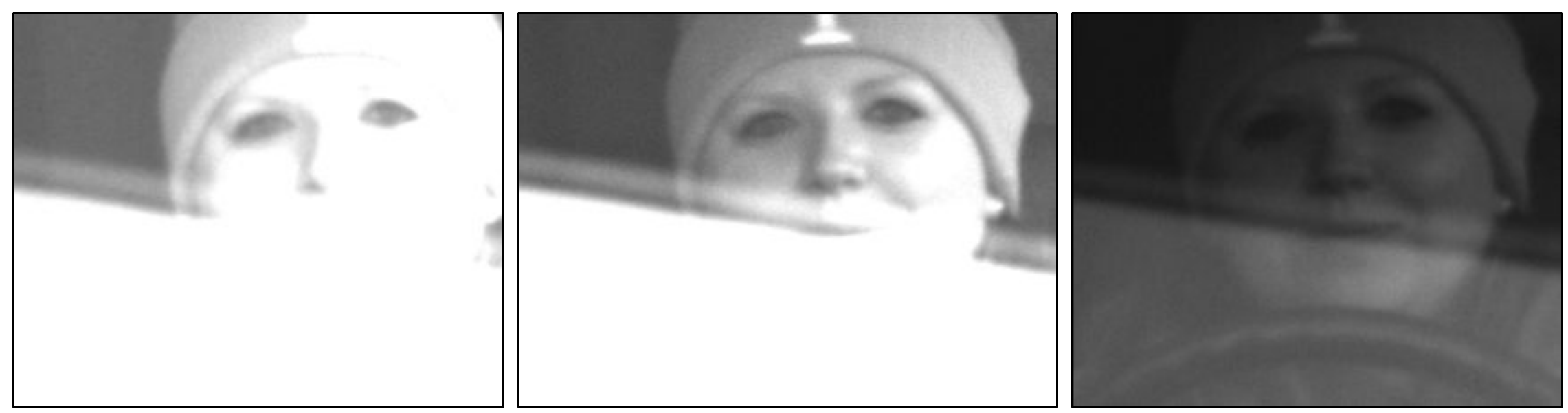}
\caption{A typical example of an HDR burst captured from the Through the Windshield System, including heavy glaring artifacts.}
\label{fig:teaser}
\end{figure}

Furthermore, windshields are reflective, which produces several unique challenges. First, it further reduces the light available to the sensors. Second, the windshield will reflect light from other sources into the sensor. If that light propagates directly from a light source, this will often cause obstructive glare (Figure \ref{fig:teaser}). Even if that light is reflected off of an object onto the windshield, the object will appear as an unwanted artifact overlaid with the desired image of the driver.

Current solutions for this problem include flashing visible lights at the driver, such as the system devised by Gatekeeper Security \cite{Gatekeeper}. This is highly undesirable, as it both distracts the driver from safe vehicle operation and has the potential to cause obstructive glare due to the windshield.

To provide the best possible input to a deep learning algorithm, a custom multi-camera imaging system was developed to specifically mitigate these hurdles while remaining non-intrusive \cite{cornett_yen_nayola_montez_johnson_baird_santos-villalobos_bolme_2019}. The system is modular in design, where each unit is composed of both an imaging system and an associated computational system, seen in Figures \ref{fig:system_camera}, \ref{fig:system_computer}, and \ref{fig:system-overview}.  The raw images captured by the system can subsequently be processed by any HDR method to ultimately provide the processed input to facial recognition software.  

This paper extends the previous work in two main ways: First, we collect and annotate a new dataset for the purpose of training HDR reconstruction models fine-tuned for our specific imaging hardware. Second, we propose an novel network architecture for HDR reconstruction by using an unrolling technique to model the Mertens-Kautz-Van Reeth HDR algorithm \cite{MertensHDR} as a neural network - we call this network the Mertens-Unrolled Network, or \textbf{MU-Net}. Finally, to analyze the end-to-end system, SOTA facial detection and recognition algorithms are used to evaluate the system's image capture and HDR reconstruction quality in its overall goal to detect and identify faces through the windshield of moving vehicles. We also compare our efficacy of MU-Net against other classical and deelply-learned HDR methods from previous works in the same recognition pipeline. 

\section{Motivation}
The applications of this system are numerous. For example, the ability to recognize drivers through the windshield has the potential to speed up security checkpoints in restricted areas. These can also be permanently deployed along roads to locate criminals and terrorists. Moreover, a more agile version of this system can be used to greatly accelerate the flow of traffic through temporary checkpoints which have been set up to catch identified criminals who have recently committed a high-profile crime or escaped from prison. Alternatively, it can be used to help locate a child who is the focus of an amber alert, which can possibly return them to safety faster.

However, contrast within the vehicle is a function of the ambient light, the shadows from the vehicle, and the tint of the windshield. We should expect that these conditions may change quickly; by the addition or removal of cloud cover, or by a passing driver with their lights on. If we have a single camera with fixed optics, then we cannot expect the image to have good contrast most of the time. Unfortunately, automatic camera gain is difficult given these situations. This is why HDR fusion techniques for our camera system are so valuable - we expect that if we take pictures from multiple cameras with varying optics, we should be able to get a high-quality image by combining them.

\section{Related Work}

There is a notable body of work in the field of HDR fusion, a review of which can be found in \cite{HDRReview}. Note the algorithms given by Mertens et. al. \cite{MertensHDR} and Debevec et. al. \cite{DebevecHDR} usually perform adequate HDR fusion under normal circumstances. However, when dealing with through-windshield environments, these methods are not designed to account for the unfortunately common cases of obstructive glare, which obscures the face within one or more images in the HDR burst.

Most external driver recognition research has been performed in the commercial sector \cite{ITS_komoto} and public sector \cite{brandom_2018}. Beyond an initial case study at Oak Ridge National Laboratory \cite{cornett_yen_nayola_montez_johnson_baird_santos-villalobos_bolme_2019}, very little open research has been published on the topic. While little work has been done in the field of driver recognition through the windshield , there has been significant related work two related fields: Vehicle occupancy detection, and glass artifact suppression. In vehicle occupancy detection \cite{hao2011occupant,morris2017sensing,fan2013front,artan2013occupancy}, algorithms aim to detect and accurately count the number of occupants in an automobile, usually for the purpose of automatically enforcing HOV and carpool lane rules. In glass artifact suppression, algorithms look to effectively remove artifacts such as reflection, refraction, and glare, that are introduced by glass into real-world images \cite{kong2013physically,eccv2018/Wieschollek,fan2017generic,arvanitopoulos2017single}. While these works are useful within their own context, they are solutions designed around separate problems and therefore sub-optimal to solve the through-windshield recognition task, due to less-stringent time or accuracy constraints on their output

Convolutional Neural Networks (CNNs) are the the current SOTA approaches for a large number of image processing tasks. Some recent successes of the neural networks can be attributed to their ability learn and optimize the behavior previously established deterministic algorithms, as is pointed out in \cite{McCannJinUnserReview}. Moreover, the technique of taking an algorithm and turning it into a neural network - for example, through ``deconstructing'' or ``Unrolling'' - has been shown to effectively incorporate domain-specific priors into a given model\cite{AdlerOktem,RIM}. This technique can provide a straightforward improvement to a problem's solution, as long as the solution in question is already acceptably effective and easily posed as a network architecture. If we have designed and initialize a network with parameters to accommodate a target algorithm, we can ensure that the original algorithm can not only be mimicked by the neural network, but further improve upon it through fine-tuning \cite{rao2016deep}. We should expect an algorithm re-formulated as a neural network to perform no worse than the original.

There is a growing body of work focused around performing HDR reconstruction using a CNN. Many of them attempt to do this using a single shot \cite{OneCamHDR, AnotherOneCamHDR} or by simply having a extreme number of weights \cite{OneCamHDR, HugeHDRNet, StupidHugeHDRNet, AnotherOneCamHDR}. For real-time, robust driver recognition, both of these are unacceptable. In the former case, we expect that a single shot will have a high probability of either misfiring or inaccurately capturing the face, due to the potential for obstructive glare. In the latter case, an extremely large model cannot be run locally, within an embedded system on-device, and expected to perform with adequate throughput for the frequency and volume of highway traffic. Previous work attempts to overcome these issues using a GAN approach \cite{cornett_yen_nayola_montez_johnson_baird_santos-villalobos_bolme_2019}, which this work improves upon. While our proposed Deep HDR Pipeline, the MU-Net, is designed and trained to work specifically with the our custom face imaging system, it's architectural design mimicking the general Merten's HDR algorithm inherently improves the method's generalizability across all domains, not just through-windshield faces.

\section{Computational Imaging System Overview}

\begin{figure}[b!]
\centering
\includegraphics[width=5.5in]{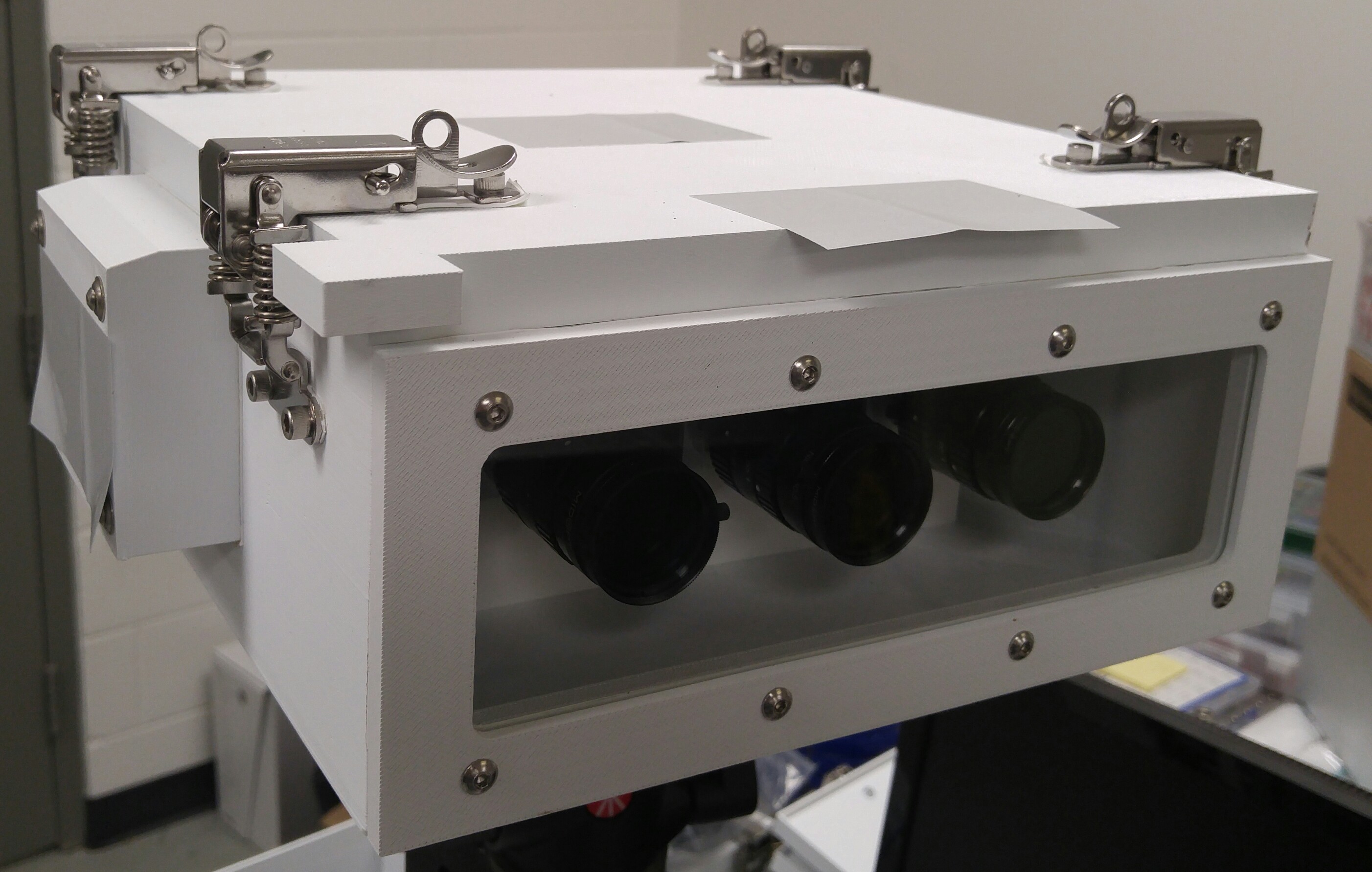} 
\caption{Computational imaging unit - weatherproof camera array.}
\label{fig:system_camera}
\end{figure}

\begin{figure}[b!]
\centering
\includegraphics[width=5.5in]{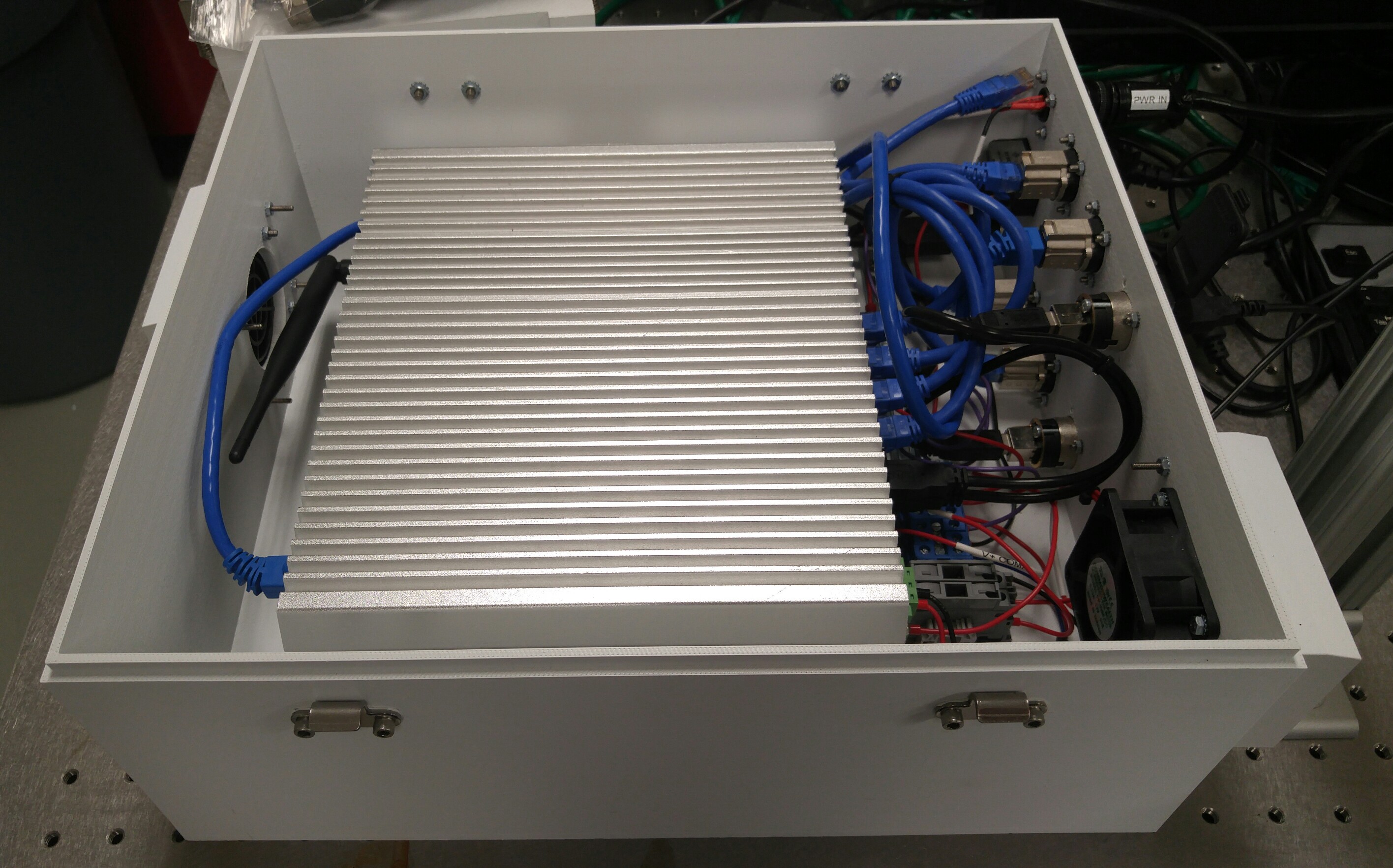}
\caption{Computational imaging unit - weatherproof ruggedized computer.}
\label{fig:system_computer}
\end{figure}

Our custom imaging system was developed to withstand harsh environments as well as produce quick results for biometric assessment in field deployment.  The overall system design is modular, where each computational unit added can perform stand-alone, or input its data into a central analysis system.  A single computational imaging unit is comprised of two primary components: weatherproof camera enclosure and a ruggedized computer.  The camera enclosure is shown in Figure~\ref{fig:system_camera} and the associated computing system can be seen in Figure~\ref{fig:system_computer}.

The camera enclosure houses an array of three Basler GigE cameras ($2048 \times 1536$ pixel matrix, 3.45 $\mu$m pixel size), mounted on a 3D-printed bracket and arranged horizontally. The cameras are linked to the associated computer unit via Power over Ethernet (PoE) connections. They are aligned to have a common focal point at 10 meters. Gain and exposure time is programmatically adjusted at the time of data acquisition. The cameras are equipped with filters as follows: First, all cameras are equipped with a 50 mm focal length and 34 mm aperture lens. Next, all cameras are equipped with a Midwest Optical Systems UVIR filter to sample only visible light. Third, Cameras \#1 and \#2 are equipped with a Midwest Optical Systems 0.9 and 0.3 ND filter, respectively. Last, cameras \#1, \#2, and \#3 are equipped with Meadowlark Optics linear polarizers (extinction ratio greater than 10,000,000:1) placed at angles of $80^\circ$, $90^\circ$, and $100^\circ$, respectively.

The ruggedized computer is also housed in a 3D-printed, weatherize enclosure.  The computer has 32GB of memory, Intel Core i7 CPU with 8 processing cores and a NVIDIA GTX 1050 Ti GPU.  The computer was chosen for its PoE ports used to interface with the camera unit as well as its GPU so that it can handle the computational burden of the implemented deep learning algorithms.

To trigger the imaging system an Optex through beam photoelectric sensor was used.  To ensure frame synchronization across multiple systems the same trigger signal was distributed to all systems.  Once triggered, an imaging unit captures 20 frame-sets (20 x 3 cameras = 60 images per frame-set).

The system software is also modular by design, and leverages the Google Remote Procedure Call (gRPC) library.  The gRPC library allows for the use of independent servers so that each task of the image acquistion and processing pipeline can be managed independently.  

\begin{figure}[b]
\centering
\includegraphics[width=5.5in]{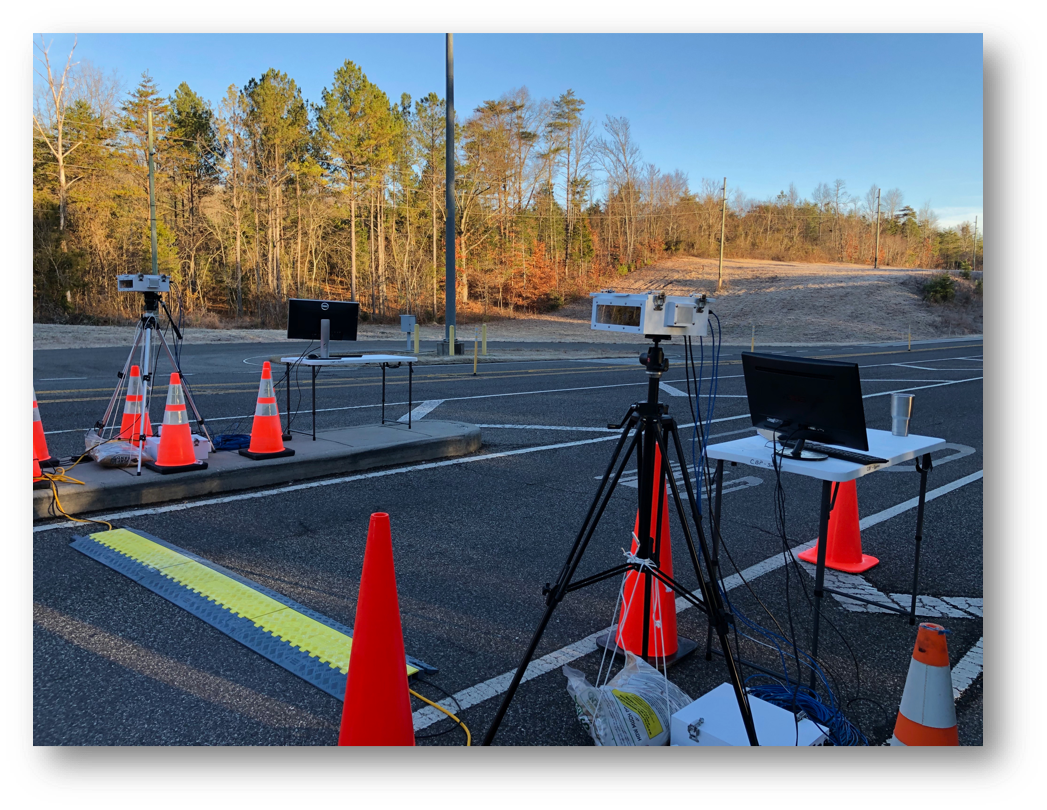}
\caption{Two-unit camera setup for dataset collection at security portal.}
\label{fig:system-overview}
\end{figure}

For the all experiments mentioned in this paper, two modular units were used - one on either side of the driving lane that test subjects would drive through.  This configuration was used to account for driver head pose, ensuring that a sufficiently frontal face shot could be captured for detection and identification tasks.  The layout of the units can be observed in Figure~\ref{fig:system-overview}.

\section{Overview of Image Pipeline}
\label{sec:imagepipeline}
Before images are acquired, the cameras are calibrated to give an initial estimate for coarse image registration; After this, we use FaRO's \cite{bolme2019faro} implementation of RetinaFace face detection on the three images. Using these detection locations, a secondary registration is performed (in the case that a face was not detected in a given image, this step is skipped). We crop the images to 600 $\times$ 600 pixels and perform fine registration using a modified form FFT-based image registration \cite{FineReg}; We modify the algorithm to multiply the images using a Kaiser window with $\beta = 4$ as to ``prefer'' smaller shifts (we expect the fine registration to be, in our case, usually between $30$ and $150$ pixels in any given direction, as we have already been coarsely registered). We utilize this method of registration to achieve accurate correspondence and alignment without the need to detect facial landmark points, which are often noisy or incorrect within our dataset. It should be noted that this method deviates from the previous work, \cite{cornett_yen_nayola_montez_johnson_baird_santos-villalobos_bolme_2019}, which uses Hamming windows to perform alignment instead. The images are then cropped to $256 \times 256$ pixels and then fed into the HDR fusion neural network outlined in Figure \ref{fig:MU-Net}. This produces a final face tile which can then be fed to a biometric template extractor.

\section{MU-Net: Mertens Unrolled Network}
Motivated by the simplicity of the Mertens HDR algorithm, we first wanted to show that the algorithm itself could be reformulated within a CNN architecture. Therefore, MU-Net was conceived by applying an ``Unrolling'', first used in \cite{UnrolledISTA}, to the Mertens algorithm. 

For completeness, a rough sketch of the Mertens HDR algorithm follows. First, for each pixel, the contrast and well-exposedness of each pixel is computed, by the formulas

\begin{equation}
\begin{split}
C_{i,j,k} = | R_{i-1,j,k} + R_{i+1,j,k} + R_{i,j-1,k} \\
     + R_{i,j+1,k} - 4 R_{i,j,k} |
\end{split}
\end{equation}
and

\begin{equation}
X_{i,j,k} = \exp\left( -\frac{1}{2} \left(\frac{R_{i,j,k} - 0.5}{\sigma}\right)^2 \right)
\end{equation}

where $R_{i,j,k} \in [0,1]$ is the $(i,j)$th pixel of the $k$th image, and $\sigma$ is some standard deviation (in the standard implementation $\sigma = 0.2$.) Concretely, $X$ is a measure of image exposure, while $C$ is a measure of contrast. We note that for a color image, there is also a measure of saturation; this network can be extended to include this saturation measure, but in our greyscale implementation it is not necessary.

After this, the $C$ and $X$ measurements are multiplied to give a measure of ``quality'' of each pixel of each image. This quality metric is then normalized, and then the images are mixed using a Laplacian Blending scheme, as in \cite{LaplacianPyramidBlending}.

\begin{figure*}
    \centering
    \includegraphics[width=5.5in]{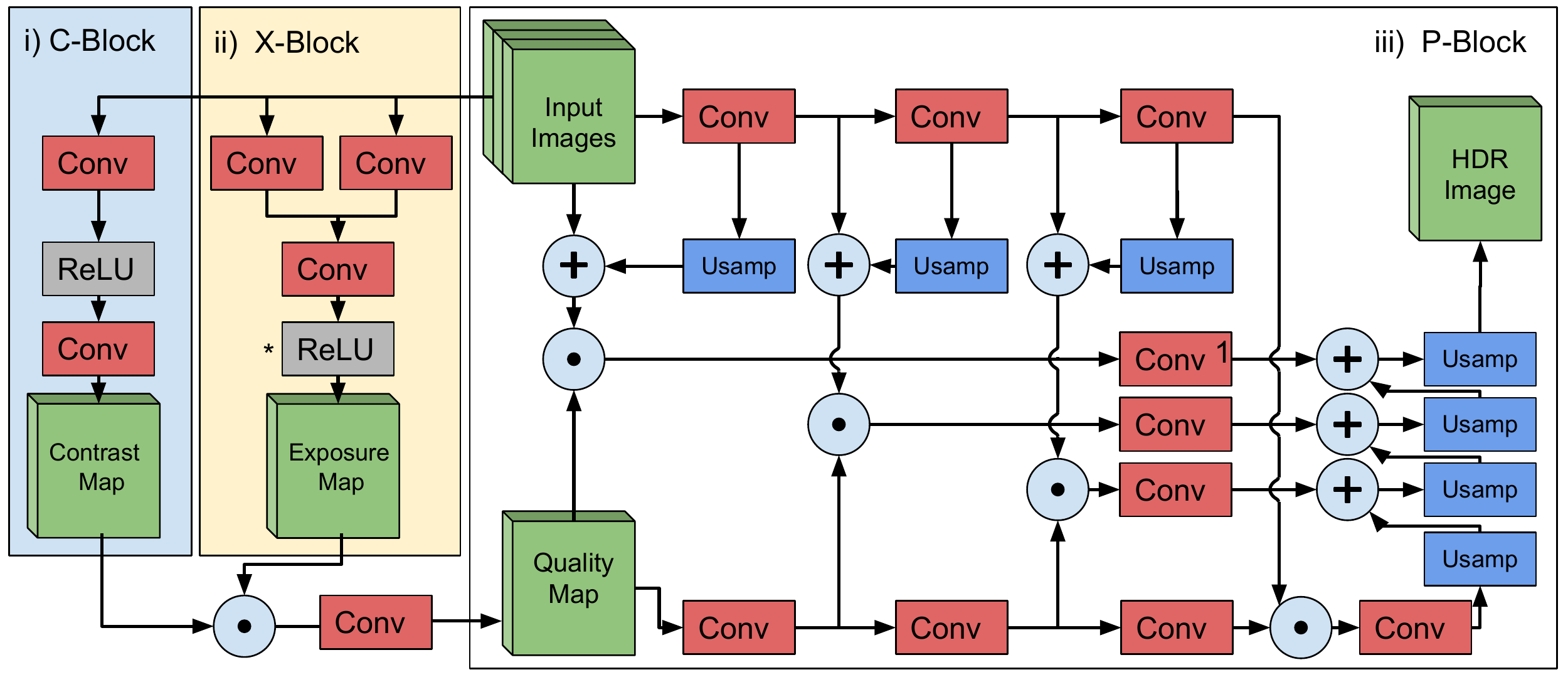}
    \caption{An overview of the MU-Net architecture. From left to right: i) The Contrast Block (C-Block), ii) The Exposure Block (X-Block), iii) the Concatenation Block (P-Block). Note that P-Block shows only 3 unrolled loops for brevity. Each convolutional and upsampling layer has a stride of 2 with 3 output channels, unless denoted by a superscript. Each conv layer utilizes tanh activation unless otherwise denoted with a ReLU block. * - contains an ELU, Negative, Exp, and Tanh activation.}
    \label{fig:MU-Net}
\end{figure*}

We determine that the way to proceed is to develop a network composed of three blocks, following the scheme above. First, a Contrast Block, or ``C-Block'', which is initialized to compute the contrast as above. Second, a Well-Exposure Block, or ``X-Block,'', which is initialized to compute the Well-Exposure as above. Third, a Pyramid Blending Block, or ``P-Block,'', which is initialized to perform Laplacian Pyramid Blending. The output of the C-Block and X-Block is multiplied together, a convolutional layer is applied, and the resulting ``Image Quality'' is fed into the P-Block along with the original images. A sketch of the neural network, as well as the blocks, are in Figure \ref{fig:MU-Net}. Note that only 3 ``steps'' of the P-Block are shown in Figure \ref{fig:MU-Net}, for brevity. In reality, our network contains 8 (that is, the network should be naturally extended so that the output of the upper-right and lower-right 2D convolutional layers are $1 \times 1$ pixel each, as each convolutional layer has a stride of 2) 

Unless denoted with a RelU block, each convolutional layer in Figure \ref{fig:MU-Net} is followed by a tanh activation. The C-Block's first convolutional layer has 6 output channels, while its second has 3 output channels, using $7 \times 7$ kernels. Each layer in the X-Block has 3 channels with $3 \times 3$ kernels. The ReLU layer that has been starred within the X-Block has an ELU, Negative, Exp, and Tanh activation. The output of the last Conv2D layer represents an exponent, whose value we want to remain negative so that our network's output never ``explodes''. The ELU and Negative force that condition on that parameter. The Conv2D layer between the P-Block has $3$ output channels and a $3 \times 3$ kernel. Every Conv2D layer in the P-block has a $7 \times 7$ kernel and a tanh activation function. All of them have 3 output channels, with the exception of the ones which are superscripted.




As we developed this generator by unrolling the Laplacian Blending scheme in the Mertens algorithm, we call it the Mertens Unrolled Network, or ``MU-Net.'' Note that the architecture is rather light-weight: our current architecture has under $45,000$ parameters.

A significant advantage to this approach is that we avoid treating our neural network like a ``black box.'' Specifically, we expect that our C-Block measures something similar to the contrast, our X-Block measures something similar to the Well-Exposedness, and that our P-Block performs something similar to Laplacian Pyramid Blending. This means that it should be possible to debug the MU-Net, and we can identify when and where our training has gone badly.

\subsection{MU-Net Discriminator: The NU-Net}
We use a Feature Pyramid Network (FPN) architecture to design our discriminator \cite{FPN}. The FPN provides a multi-scale framework for learning a feature to detect real and fake images at different frequency scales. If we take the maximum element in each channel (that is, if we maxpool down to $1 \times 1$ pixel), this should serve as an indicator of how much each feature exists within the input image. This pooled result is then feed that a classifier layer with a single dimension output. We decided to downsample more aggressively than in the MU-Net to reduce the number of parameters required; it was found that it was easy for the architecture to overtrain if it was given more parameters, so we opt to use a smaller network. Moreover, we feed the contrast well-exposure maps from the C-Block and X-Block into the discriminator, as these are pertinent features to generate correctly when producing a fused HDR image.

\begin{figure}
    \centering
    \includegraphics[width=5.5in]{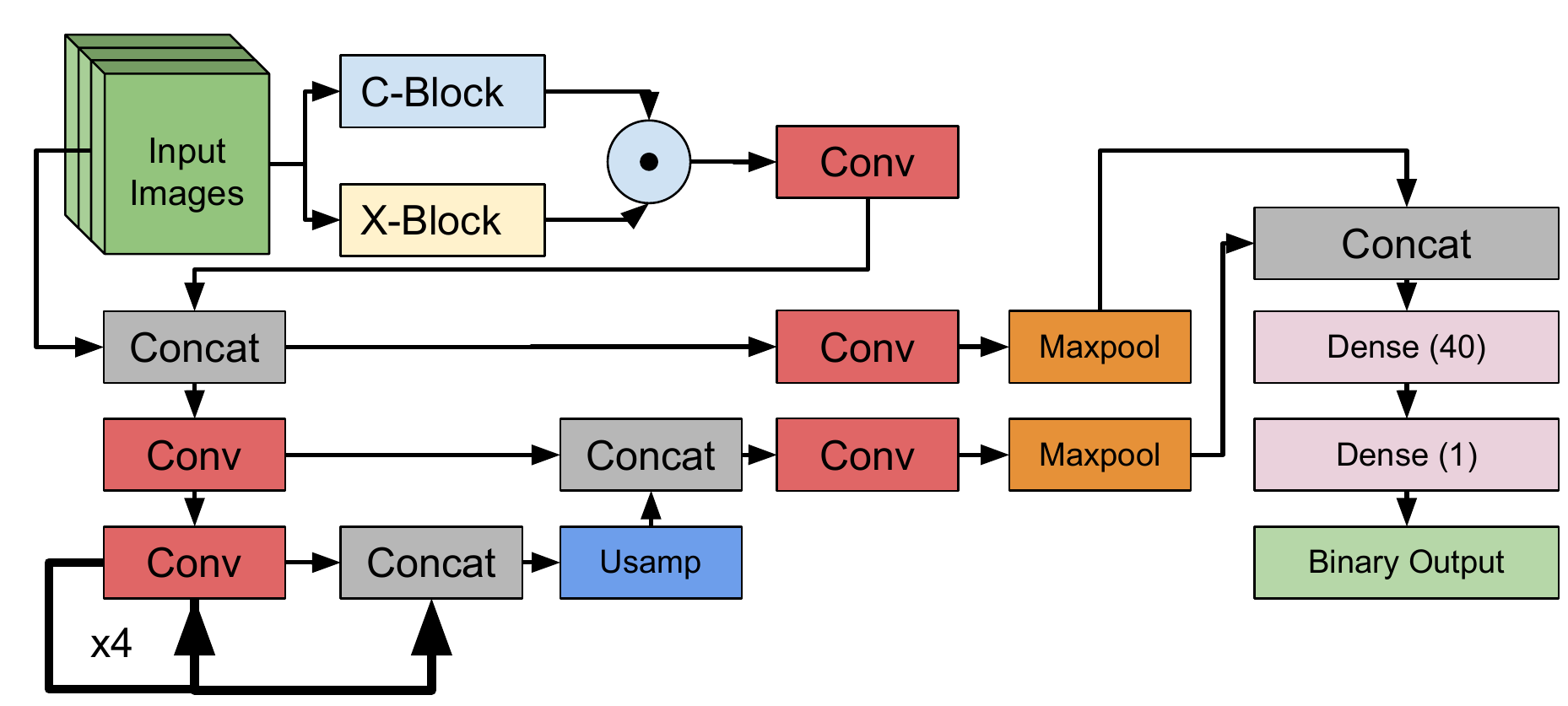}
    \caption{NU-Net (Discriminator) Architecture.} 
    \label{fig:NU_Net}
\end{figure}

The architecture for NU-Net is shown in Figure \ref{fig:NU_Net}. The upper half of the diagram is the same as in the MU-Net, utilizing the C- and X-Block. The leftmost Conv2D layers all have a $7 \times 7$ kernel, except for the last, which has a $4 \times 4$ kernel, due to reduced resolution. They have 4,8,16, and 32 output channels, from top to bottom, respectively. The upsampling Conv2D layers have 16 output channels and a $7 \times 7$ kernel, except the first, which has a $3 \times 3$ kernel due to reduced resolution. The rightmost Conv2D layers all have $8$ output channels, with $5 \times 5$ kernels.



As we developed this discriminator by doing something other than unrolling, we call it the Not Unrolled Network, or \textbf{``NU-Net.''} Both the MU-Net and NU-Net were built in Keras, using the Tensorflow backend.

\subsection{Training}
\label{sec:training}
We were able to take a number of photos at the security gate of a restricted access vehicle portal. From this set of photos, we were able to capture $1957$ triples of images which were of faces, as well as $2371$ images of faces which were of sufficiently high quality for training the discriminator. Of this, we used $1440$ of the triples and face tiles for training, $160$ for validation, and the rest as a testing set.

We determined that the use of Batch Normalization layers to overcome the exploding gradients problem was not a good solution; naive approaches to Batch Normalization eliminate the advantages of our initialization, due to our initialization passing values of nonzero mean and nonunit variance between layers. We opted to perform $L1$ normalization on the layers, with $\lambda = 0.001$, along with gradient clipping, with our clipnorm value being $0.1$. Moreover, we determined that pretraining the adversary for $10$ epochs would improve the initial training, as our generator had a ``head-start''  due to its good initialization. We used ADAM to optimize, with an initial learning rate of $0.005$, a learning rate decay of $0.000001$, and a value of epsilon equal to $0.001$; we set all other parameters to the defaults.

During training, we noticed that the output of our GAN began to degenerate once the adversary began to learn. It seems that this is due to the generator learning a ``cheap trick'' that fools the adversary, but does not produce realistic looking images. This was solved by taking the training result before degeneration; we save output images after each epoch, and take the result that looks qualitatively best. Although we trained for $100$ epochs, we determined that the result of training after the $71$st epoch provided the best aesthetic quality.

\section{Experimental Evaluation and Results}
This section details our effort to evaluate and analyze the effectiveness of the MU-Net HDR algorithm for the purpose of through-windshield facial recognition.
\begin{figure}[t]
    \centering
    \includegraphics[width=5.5in]{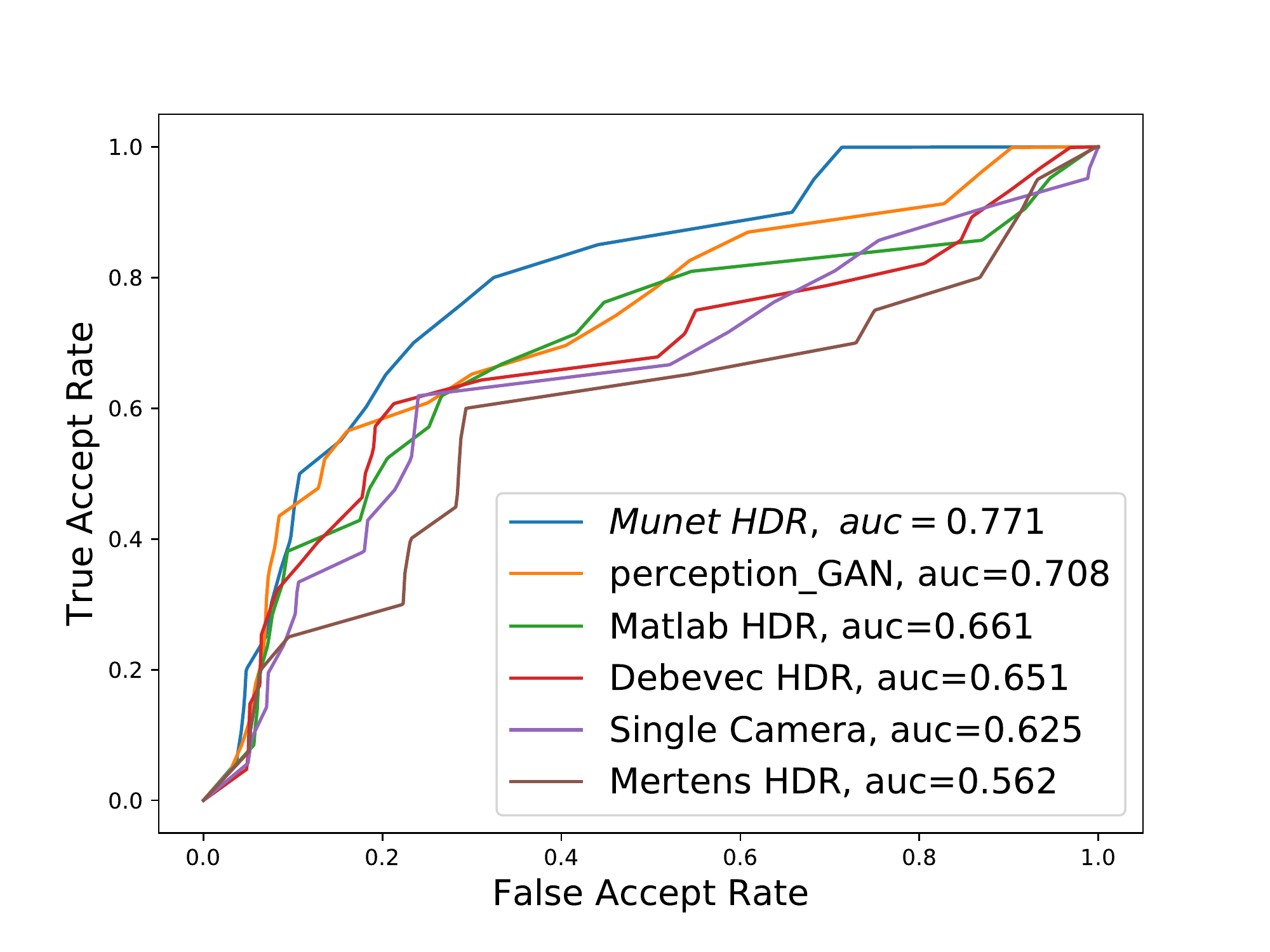}
    \caption{ROC performance of MU-Net compared to other HDR Methods.}
    \label{fig:roc}
\end{figure}

\subsection{The Driver Dataset}
\label{sec:dataset}

To evaluate MU-Net on the Image pipeline from Section \ref{sec:imagepipeline}, we capture a new dataset of known drivers, disjoint from the training data captured in Section \ref{sec:training}. The dataset was captured at a facility security portal using live driving conditions over the course of two days.  For this experiment there were 22 participants enrolled as gallery subjects for the recognition set.  These participants entered the facility through the security portal under normal driving conditions.  The images captured during this experiment represented the scene illumination changes that occur through sunrise to sundown.  There were 471 vehicle triggered events that produced 56,520 raw images.  For each of these events that involved an enrolled participant, each of the three raw image frame-sets were fused via each of the evaluated HDR methods to be used as probes by the face identification systems.

\subsection{Baseline Methods}
\label{sec:baselines}

Classical HDR fusion can be performed in a number of ways, usually combining images taken at different radiance ranges and fuses them into a single radiance image.  This method allows the fused image to display a range of radiance levels that could not be captured by a single image.  For our application, HDR imaging provides a means to overcome the artifacts introduced into face images from windshield glare and reflection.

A brief summary of all methods we will compare to our novel approach is laid out below:

\begin{itemize}
    \item \textbf{Matlab HDR}\cite{reinhard2010high} is the standard Matlab implementation for HDR fusion.  It is used as a standard comparison against the other methods.
    \item \textbf{Mertens-Kautz-Van HDR (Mertens)}\cite{MertensHDR} is based on the Mertens algorithm utilizes contrast and exposure estimates to fuse the three raw frames for detection and recognition. 
    \item \textbf{Debevec HDR}\cite{DebevecHDR} is based on Debevec radiance mapping with Durand tone mapping.  This method is also a physics-based implementation of HDR.
    \item \textbf{PeceptionGAN}\cite{cornett_yen_nayola_montez_johnson_baird_santos-villalobos_bolme_2019} is a GAN network trained specifically for visual perception and was the initial design for the system used in \cite{cornett_yen_nayola_montez_johnson_baird_santos-villalobos_bolme_2019}.
\end{itemize}

\begin{figure*}
    \centering
    \includegraphics[width=5.5in]{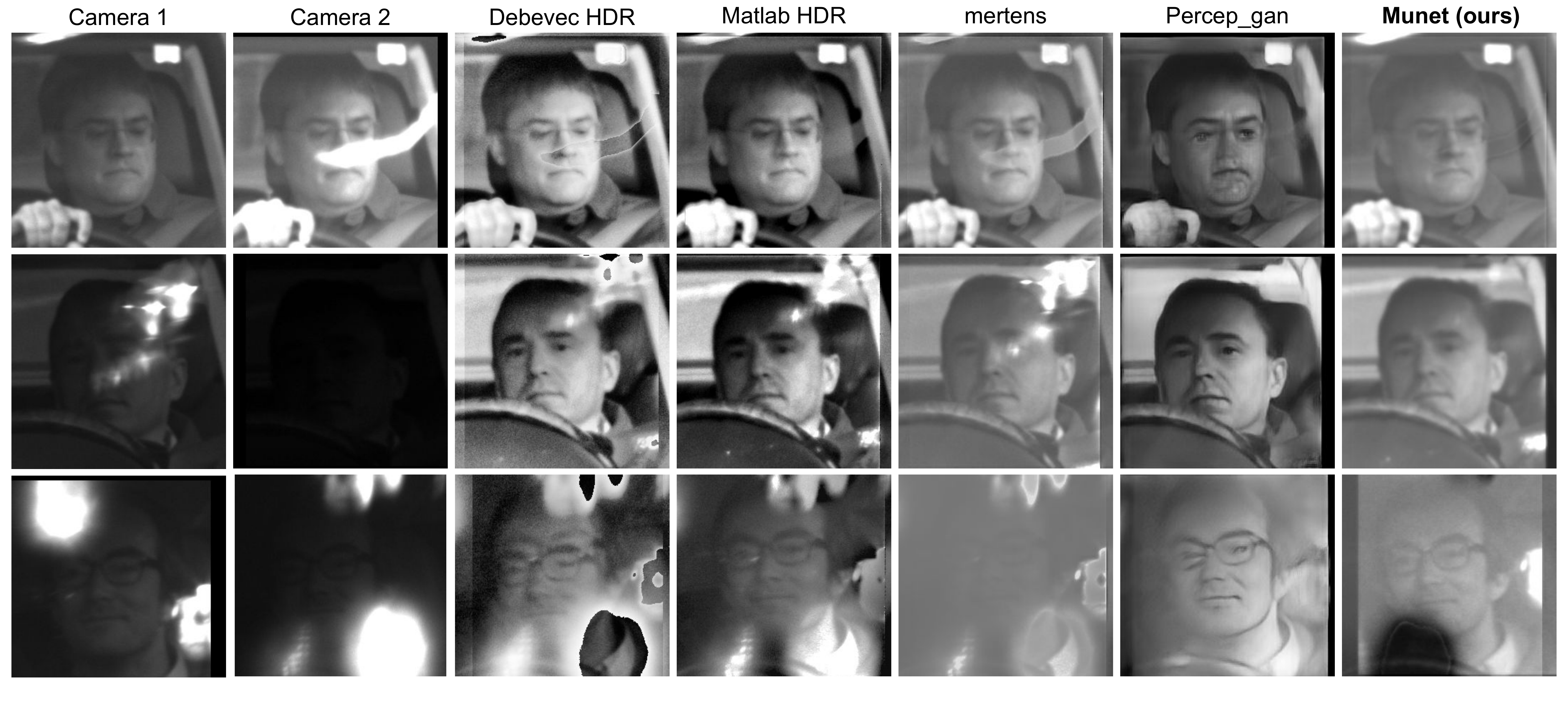}
    \caption{Sampled images from cameras (top), input to HDR (middle), Mertens HDR reconstruction (bottom left), and MU-Net HDR reconstruction (bottom right).}
    \label{fig:qualitative_matrix}
\end{figure*}

%

%
To analyze the efficacy of MU-Net, we performed a verification study on the dataset introduced in Section \ref{sec:dataset}, comparing performance against the 5 baselines described in Section \ref{sec:baselines}. We also include the baseline \textit{Single Camera}, which uses a randomly-selected, non-processed image from each HDR burst. For each method, we detect faces using RetinaFace \cite{deng2019retinaface} and subsequently extract feature templates using Additive Angular Margin Loss for Deep Face Recognition (ArcFace)\cite{deng2018arcface}. Because our driver dataset is relatively small and was not used for network training, we report results over the entire set of subjects. Figure \ref{fig:roc} Shows the final results of all 6 methods. The MU-Net provides the best performance out of all those evaluated, improving upon the PerceptionGAN by almost $7\%$. As can be seen by the AUC's of most other methods, this problem is inherently difficult to solve, and classical HDR methods do not provide high enough quality reconstructions for accurate face recognition. Figure \ref{fig:qualitative_matrix} shows that classical methods like Debevec and Merten's are not robust towards artifacts introduced by windshield glare and reflection, which significantly impairs verification performance.

Quantitative and Qualitative analysis of the MU-Net pipeline revealed two key features of the system: First, MU-Net is better able to discriminate areas of poor quality, including glare and reflection artifacts, than the original Mertens or Debevec HDR methods, as seen in Figure \ref{fig:qualitative_matrix} columns 3,4, and 5. Second, MU-Net provides output reconstructions that are more stable than that of PerceptionGan. Because the underlying architecture of MU-Net was designed and initialized from a state that already performed classical HDR, the network overfits our data less, resulting in fewer GAN-like artifacts (see Figure \ref{fig:qualitative_matrix} columns 6 and 7). We see in general that MU-Net consistently outputs a more stable image free from windshield artifacts than its competing baselines.

While MU-Net improves upon previous methods for through-windshield recognition, there is still significant work to be done to achieve a highly reliable system. For instance, MU-Net can suffer when running in low-light conditions. This highlights the need for more uniform data collection, ideally over the course of an entire day. Moreover, the MU-Net was trained on faces - it does not appear to work properly on images that are not faces.

\section{Conclusion and Discussion}

The MU-Net architecture described in this paper represents a significant step forward in the use of neural network architectures for HDR fusion for face recognition through the windshield. Using this method we are able to obtain higher quality images than classical fusion methods while maintaining a relatively low computational overhead. While other works have performed HDR using deep learning, our method is unique in that it is designed on top of the original Merten's HDR algorithm, and has a relatively low number of parameters in contrast to other comparative algorithms. The MU-Net shows that we are able to successfully incorporate exposure and contrast quality priors into our deep learning architecture, allowing for faster training, more stable fusion results that introduce significantly fewer GAN-related generative artifacts.

While the MU-Net is a major improvement over previous methods, significant work must be done to further enhance facial recognition in the adverse conditions of through-windshield imaging. We believe it would be beneficial to train the network stacked end-to-end with a face recognition algorithm, eliminating the need for training with an ad-hoc discriminator, and optimizing the HDR fusion expressly for the task of facial recognition.

Efforts are currently being made to expand the custom dataset of through the windshield images.  This expansion would include more enrolled subjects, a more expansive training set, and a wider array of weather and illumination conditions.  Additionally, the custom imaging system is being redesigned with better optics, utilizing polarized filter arrays to mitigate glare for different angles of incidence. 

\acknowledgments 

This research was supported in part by an appointment to the Oak Ridge National Laboratory Post-Bachelor’s Research Associate Program, sponsored by the U.S. Department of Energy and administered by the Oak Ridge Institute for Science and Education, the U.S. Department of Energy, Office of Science, Office of Workforce Development for Teachers and Scientists (WDTS) under the Science Undergraduate Laboratory Internships Program (SULI), and the Texas A\&M University-Kingsville, Office of University Programs, Science and Technology Directorate, Department of Homeland Security Grant Award \# 2012-ST -062-000054.

\bibliography{report} 
\bibliographystyle{spiebib} 

\end{document}